%% file: main.tex
\documentclass[10pt,twocolumn,letterpaper]{article}

\usepackage{cvpr}              
\input{preamble}
\definecolor{cvprblue}{rgb}{0.21,0.49,0.74}
\usepackage[pagebackref,breaklinks,colorlinks,allcolors=cvprblue]{hyperref}

\def\paperID{*****} 
\def\confName{CVPR}
\def\confYear{2026}

\title{Prompt Relay: Inference-Time Temporal Control for Multi-Event Video Generation}

\author{Gordon Chen \quad Ziqi Huang \quad Ziwei Liu\\[5pt]
S-Lab, Nanyang Technological University\\[10pt]
\url{https://gordonchen19.github.io/Prompt-Relay/}
}

\begin{document}

\twocolumn[{%
            \renewcommand\twocolumn[1][]{#1}%
            \maketitle
            \begin{center}
                \centering
                \includegraphics[width=0.95\textwidth]{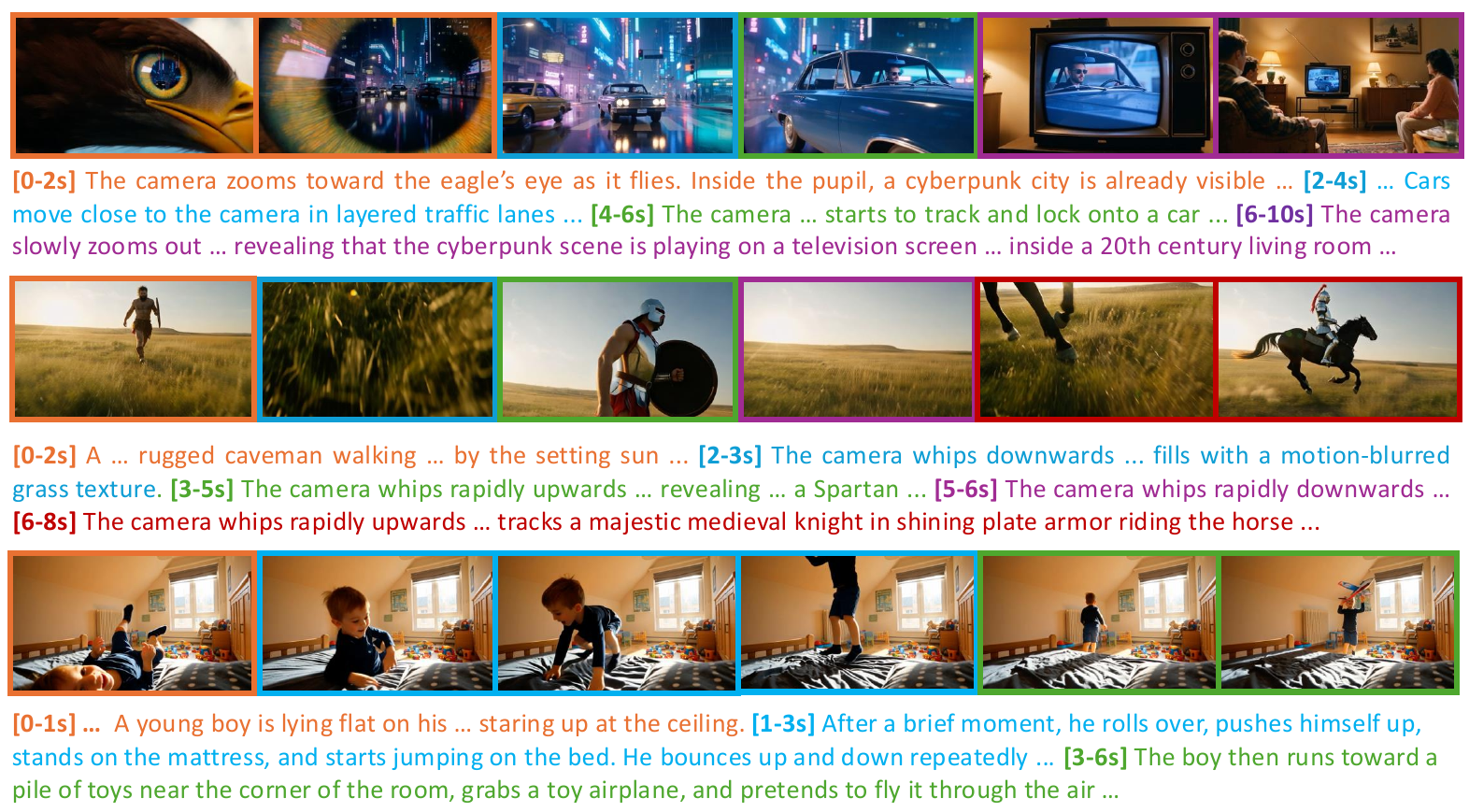}
                \captionof{figure}{\textbf{Prompt Relay} is an inference-time, training-free, plug-and-play method for enabling fine-grained temporal control by routing each textual prompt to its intended time segment, allowing multiple events to occur in the correct order without semantic interference.}
                \label{teaser}
            \end{center}%
        }]

\input{sec/1_intro}
{
    \small
    \bibliographystyle{ieeenat_fullname}
    \bibliography{main}
}


\end{document}

%% file: sec/1_intro.tex
\begin{figure*}[t]
    \centering
    \includegraphics[width=0.99\linewidth]{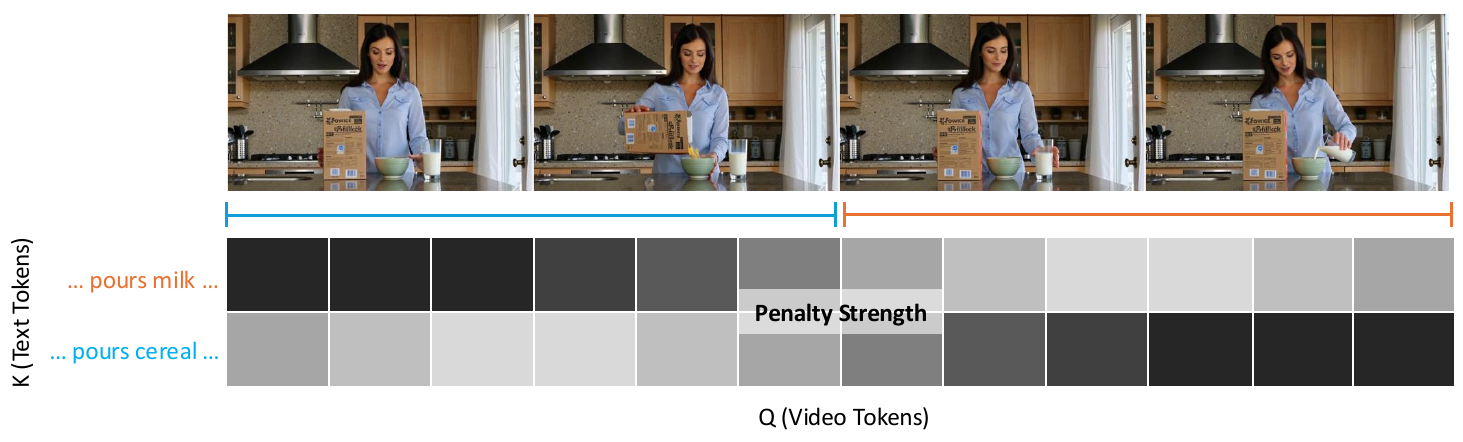}
    \caption{\textbf{Temporal Cross-Attention Routing.} 
    Each textual prompt is associated with a specific temporal segment of the video. The attention penalty varies smoothly across time, allowing video tokens to attend strongly to their corresponding prompt within the assigned interval while suppressing attention to temporally irrelevant prompts. This enables multiple events (e.g., pouring cereal followed by pouring milk) to occur in the correct order without semantic interference.
}
\label{fig:method}
\end{figure*}

\begin{abstract}
Video diffusion models have achieved remarkable progress in generating high-quality videos. However, these models struggle to represent the temporal succession of multiple events in real-world videos and lack explicit mechanisms to control when semantic concepts appear, how long they persist, and the order in which multiple events occur. Such control is especially important for movie-grade video synthesis, where coherent storytelling depends on precise timing, duration, and transitions between events. When using a single paragraph-style prompt to describe a sequence of complex events, models often exhibit semantic entanglement, where concepts intended for different moments in the video bleed into one another, resulting in poor text-video alignment. To address these limitations, we propose Prompt Relay, an inference-time, plug-and-play method to enable fine-grained temporal control in multi-event video generation, requiring no architectural modifications and no additional computational overhead. Prompt Relay introduces a penalty into the cross-attention mechanism, so that each temporal segment attends only to its assigned prompt, allowing the model to represent one semantic concept at a time and thereby improving temporal prompt alignment, reducing semantic interference, and enhancing visual quality.
\end{abstract}

\section{Introduction}
\label{sec:intro}
Recent advances in video diffusion models have enabled the generation of high-quality videos conditioned on textual prompts, achieving impressive visual fidelity and motion coherence \citep{yang2024cogvideox, wan2025wan, Veo3.1, Kling2.6, Sora}. Despite this progress, existing models are optimized for single-event generation and offer no mechanism for explicit temporal control - users cannot specify when an event occurs, how long it persists for and how multiple events are ordered. As a result, modeling movie-grade videos composed of a succession of events, actions, or camera motions, each occurring within a specific segment of the video and in a specific order, remains challenging. This limitation stems from the lack of temporal awareness in the cross-attention mechanism: by conditioning every frame of the video on the entire prompt simultaneously, the model treats a multi-event prompt as global context rather than a temporally structured sequence, causing semantic concepts intended for different moments to bleed into one another, degrading text-video alignment. 

Recent works have begun to address temporal controllability in video generation \citep{wu2025mind, oh2024mevg, cai2025ditctrl, xu2026switchcraft, zhang2026tsattn, yang2025longlive}. One line of work \citep{yang2025longlive, wu2025mind} finetunes the backbone model with temporally grounded supervision. However, these methods require large amounts of annotated data, training and shifts the pre-trained model's distribution. Inference-time attention control methods \citep{cai2025ditctrl, zhang2026tsattn, xu2026switchcraft} avoid training altogether, but impose structural constraints on the attention mechanism that limit their generality and can introduce visual artifacts at segment boundaries.

In this paper, we propose Prompt Relay, a simple and elegant attention-level routing mechanism for fine-grained temporal control and multi-event video generation. Prompt Relay operates entirely at inference time and is plug-and-play compatible with existing video diffusion backbones. Prompt Relay requires no computational overhead and no architectural modifications. Our main contributions are as follows:

\begin{itemize}
    \item We propose Prompt Relay, a test-time, plug-and-play method for fine-grained temporal control in video generation with no computational overhead.

    \item We propose a Boundary-Attention decay mechanism, a soft Gaussian penalty on cross-attention logits that smoothly suppressess semantic interference across segment boundaries.

    \item We demonstrate that Prompt Relay substantially improves temporal prompt alignment, reduces semantic interference and enhances visual quality.
\end{itemize}

\section{Related Works}



\subsection{Controllable Video Generation}

\noindent Video generation has seen rapid progress in recent years, with applications spanning motion control \citep{wang2023videocomposer, burgert2025go, wang2024motionctrl, abdal2025dynamic,  wang2025motion}, viewpoint control \citep{ren2025gen3c, he2024cameractrl,bai2025recammaster}, identity control \citep{hu2025hunyuancustom,zhong2025concat,liu2025phantom} and editing \cite{liu2024video, bian2025videopainter}. However, most models remain limited in the ability to generate coherent multi-event videos. Because the attention mechanism allows every pixel to attend to every prompt token, models struggle to associate semantic concepts with their intended temporal intervals, leading to temporal misalignment and semantic entanglement. This challenge motivates us to provide explicit temporal control at inference time.

\subsection{Attention-Based Control in Diffusion Models} 

\noindent Attention manipulation has emerged as a key mechanism for controllable diffusion generation. Prior work has explored attention for spatial \citep{hertz2022prompt,chen2025stencil, chefer2023attend, cao2023masactrl, zhang2023adding}, identity \citep{zhou2024storydiffusion, cai2025mixture} and motion control \citep{liu2024video, wang2025motion, meral2024motionflow}. In contrast, attention-based temporal control remains largely underexplored.

\subsection{Multi-Event Video Generation} 

\noindent A notable approach to temporal modeling for multi-event video generation is MinT \citep{wu2025mind}, which introduces a trainable temporal cross-attention module that binds event descriptions to predefined time intervals, but requires additional training, architectural modifications, and temporally annotated data. MEVG \citep{oh2024mevg} generates each event clip sequentially, conditioning on the last frame of the previous clip via latent inversion to maintain visual continuity. However, this autoregressive design causes error accumulation across segments and produces abrupt transitions when consecutive events are semantically dissimilar. DiTCtrl \citep{cai2025ditctrl} proposes mask-guided KV-sharing within MM-DiT's 3D full-attention, enabling prompt-specific semantic control without training. However, the binary attention masks derived from the attention map introduce hard boundaries that can cause background inconsistencies and unnatural transitions. TS-Attn \citep{zhang2026tsattn} and SwitchCraft \citep{xu2026switchcraft} instead modulate cross-attention by identifying motion-relevant tokens, TS-Attn via a subject semantic layout, and SwitchCraft via event-specific anchor tokens. Both methods therefore assume the presence of a dominant foreground subject in each event and struggle with scene-level changes or events where no single entity dominates the frame.

\begin{figure}[t]
    \centering
    \includegraphics[width=0.99\linewidth]{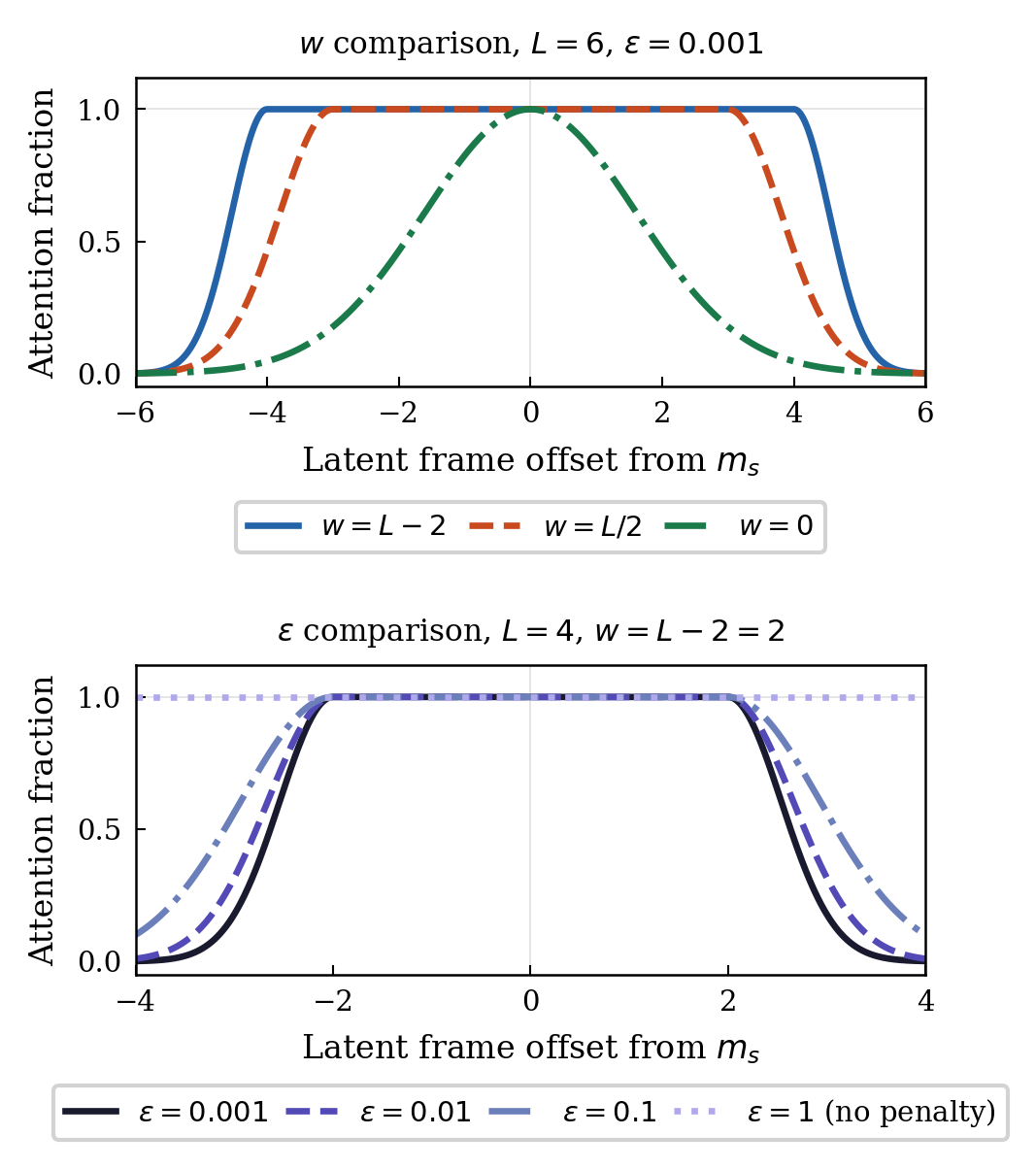}
    \caption{\textbf{Ablation Study of the Temporal Penalty Function.} 
    The curves show the attention fraction retained between a query token and the prompt tokens of a given segment, as a function of the query's latent frame offset from that segment's midpoint $m_s$, after applying the penalty $\exp(-C(i,j))$. (Top) Effect of the window parameter $w$. $w=L-2$ preserves full attention within the segment and only suppresses attention near the segment boundaries. (Bottom) Effect of the decay threshold $\epsilon$. Smaller values enforce stronger attenuation outside the 'free-attention' window; however, we find that the choice among small values has negligible perceptual impact. We adopt $\epsilon=0.1$ as our default.}
    \label{fig:attention_curves}
\end{figure}

\section{Prompt Relay}

Given a sequence of temporally-constrained text prompts 
$\{(p_s, t_s^{\text{start}}, t_s^{\text{end}})\}_{s=1}^{N}$,
our goal is to generate a video such that each arbitrary prompt $p_s$ is realized within its designated temporal interval $[t_s^{\text{start}}, t_s^{\text{end}}]$. The generated video should preserve global coherence while ensuring that each prompt influences only its assigned temporal region.

\subsection{Preliminaries}

Cross-attention is a mechanism that enables a diffusion model to incorporate external conditioning information, such as text prompts, into the generation process. Given a latent representation at diffusion step $t$, denoted as $\phi(z_t)$, and a set of conditioning embeddings $\psi(P)$ derived from an input prompt $P$, cross-attention computes interactions between the two through learned projections.
\begin{equation}
\mathrm{Attn}(\phi(z_t), \psi(P)) =
\mathrm{Softmax}\!\left( \frac{QK^{\top}}{\sqrt{d}} \right) V,
\end{equation}
where $Q = \ell_Q \phi(z_t)$ are query vectors derived from latent features, $K = \ell_K \psi(P)$ and $V = \ell_V \psi(P)$ are key and value vectors projected from the conditioning embeddings, and $d$ denotes the projection dimensionality. Each attention weight reflects how strongly a latent query attends to a particular conditioning token. Through this operation, semantic information from the conditioning input is selectively injected into the latent representation, allowing different queries to respond to different aspects of the prompt. However, because attention is computed globally over all conditioning tokens, multiple semantic concepts may compete for influence over the same latent queries. When these concepts correspond to different temporal regions, unrestricted attention can lead to interference between instructions.

\begin{figure*}[t]
    \centering
    \includegraphics[width=0.9\linewidth]{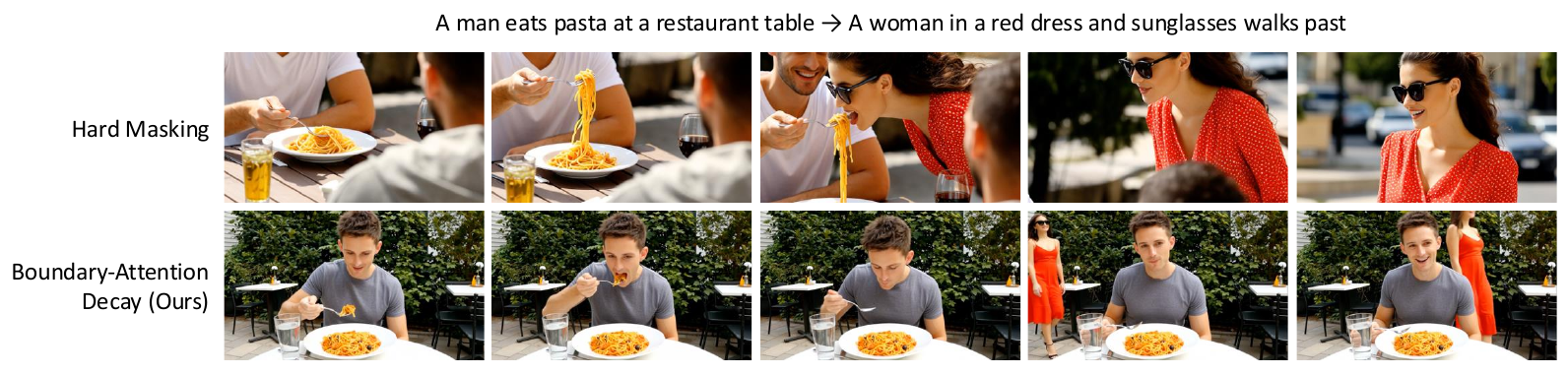}
    \caption{\textbf{Hard Masking vs Boundary-Attention Decay.} 
    Hard masking enforces an abrupt semantic switch in cross-attention at segment boundaries while self-attention remains continuous across the segments. This creates a discontinuity at the boundary, forcing the model to reconcile conflicting signals (Woman eats the pasta instead of the man). Boundary-attention decay avoids this conflict by smoothly co-activating both neighboring prompts near the boundary, giving the model a gradual handoff region in which the transition can be planned jointly before being committed to in the visual representation.}
    \label{fig:Attention_Decay}
\end{figure*}

\subsection{Temporal Prompt Routing}

In order to enforce the association between each prompt $p_s$ and its assigned temporal interval $[t_s^{\text{start}}, t_s^{\text{end}}]$, we introduce a penalty term $C(Q, K)$ into the cross-attention logits:
\begin{equation}
\mathrm{Attn}(\phi(z_t), \psi(P)) =
\mathrm{softmax}\left(\frac{QK^\top}{\sqrt{d}} - C(Q, K)\right)V .
\end{equation}
The role of $C(Q, K)$ is to suppress the attention between key and query tokens whenever they do not belong to the same interval $[t_s^{\text{start}}, t_s^{\text{end}}]$. This allows each prompt to guide generation only within its intended segment, without leaking semantic concepts into other parts of the video. For any arbitrary query token indexed by $i$ and any key token j belonging to $p_s$, the penalty is defined as:

\begin{gather}
C(i, j) = \frac{\mathrm{ReLU}(|f(i) - m_s| - w)^2}{2\sigma^2}, \nonumber \\
m_s = \frac{t_s^{\text{start}} + t_s^{\text{end}}}{2}.
\end{gather}

Here, $f(i)$ denotes the latent frame index associated with query token i, and $m_s$ denotes the midpoint of the corresponding temporal segment. The parameter $w$ defines a local window around the segment midpoint within which no penalty is applied, while $\sigma$ controls the rate at which attention decays outside this window. Query tokens within the window incur zero penalty and can attend freely to their associated prompt tokens. Beyond this region, attention is smoothly attenuated as a function of the temporal distance between the query and the segment midpoint. We demonstrate in Fig. \ref{fig:attention_curves}, that $w=L-2$ achieves the best balance between temporal isolation and intra-segment fidelity.

We compare our approach to hard masking in Fig. \ref{fig:Attention_Decay}. Hard masking sets  $C(i,j)=-\infty$ for all query-key pairs where $f(i)\notin[t_s^{\text{start}}, t_s^{\text{end}}]$ and $j$ belongs to prompt $p_s$ (i.e. a query either attends fully to a prompt or is completely blocked from it). This enforces a sudden switch between prompts at segment boundaries. While hard masking eliminates cross-segment semantic interference, it creates a discontinuity at the boundary: cross-attention switches abruptly to the new prompt while self-attention remains anchored to the previous segment's visual structure, forcing the model to reconcile conflicting signals. Boundary-attention decay avoids this conflict by smoothly co-activating both neighboring prompts near the boundary, giving the model a gradual handoff region in which the transition can be planned jointly before being committed to in the visual representation.

\subsection{Boundary-Attention Decay}
\label{app:sigma_derivation}

To suppress semantic interference across temporal segments, attention between queries near segment boundaries and prompt tokens from neighboring segments should be negligible. We therefore choose the decay parameter $\sigma$ so that the attention prior sufficiently decreases near segment endpoints. Since our penalty subtracts \(C(i,j)\) from the logits, it applies a multiplicative factor \(\exp\!\big(-C(i,j)\big)\) to the unnormalized attention scores before softmax. This prior is $1$ inside the ``free-attention'' window and decays toward the segment boundaries. Let the endpoint distance from the segment midpoint be $L=|f(i)-m_s|$. We choose $\sigma$ such that the prior reaches a small value $\epsilon$ at the endpoints:

\begin{equation}
\exp\!\left(-\frac{(L-w)^2}{2\sigma^2}\right) = \epsilon
\;\Rightarrow\;
\sigma = \frac{L-w}{\sqrt{2\ln(1/\epsilon)}}.
\end{equation}
This formulation ensures smooth transitions between neighboring prompts while preventing destructive interference across segments. As a result, each textual instruction primarily influences its intended temporal region, allowing the model to focus on one semantic concept at a time while maintaining global temporal coherence.

\begin{figure*}[t]
\centering
    \includegraphics[width=0.99\linewidth]{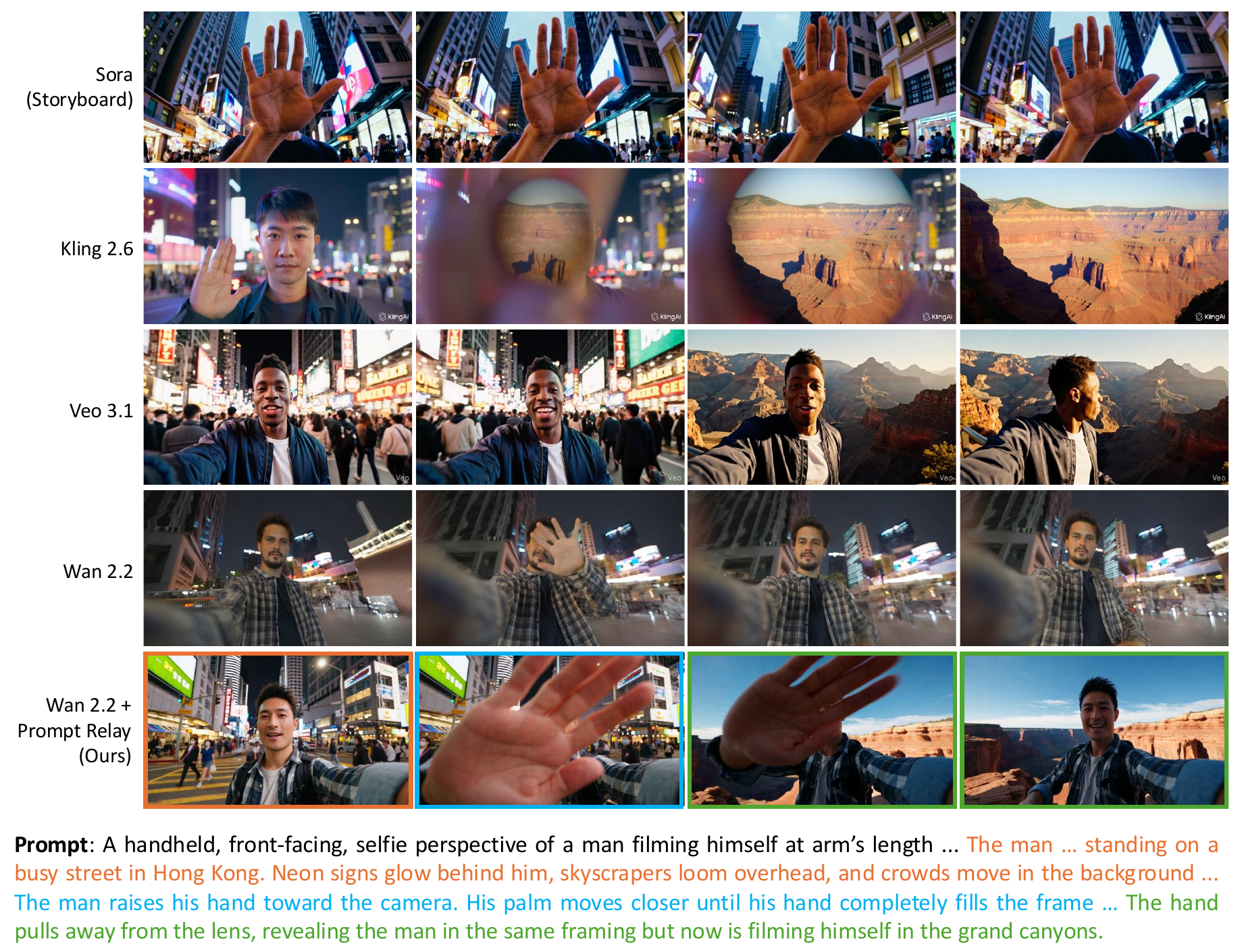}
    \caption{\textbf{Qualitative Comparison.} Given a multi-event prompt describing a deliberate scene transition, Prompt Relay preserves correct temporal structure, ensuring that each semantic instruction influences only its intended segment while maintaining global visual coherence.}
    \label{fig:qualitative_results}
\end{figure*} 

\section{Experiments}
\subsection{Experimental Setup}

We apply Prompt Relay on top of the state-of-the-art pretrained video generation model Wan2.2-T2V-A14B. To demonstrate the limitations of existing video generators in handling multi-event prompts, we test several other models, including Sora Storyboard \citep{Sora}, Veo 3.1\citep{Veo3.1}, Wan 2.2\citep{Wan2.2}, and Kling 2.6\citep{Kling2.6}. We set $\epsilon=0.1$ across all experiments. Setting $w = L - 2$ reduces $\sigma$ to a constant. In addition to selectively routing local prompts to their assigned temporal segments, we include a global prompt that conditions the entire video and provides persistent context.

\subsection{Evaluation Metrics}
Existing quantitative metrics test visual fidelity or global text-video alignment, but fail to capture temporal semantics or transition quality, properties that are inherently perceptual.  Hence, we conduct a human preference study to evaluate multi-event video generation along three dimensions: 

\begin{itemize}
    \item \textbf{Temporal Prompt Alignment:} Whether each prompt is realized in its intended temporal interval.
    \item \textbf{Transition Naturalness:} The perceptual smoothness of transitions between consecutive events, including the absence of abrupt cuts, flickering, or unnatural morphing at segment boundaries.
    \item \textbf{Visual Quality:} Overall perceptual fidelity of the generated video, including sharpness, temporal consistency, and absence of visual artifacts.
\end{itemize}

We construct 20 diverse multi-event test scenarios, covering a wide range of settings including explicit scene transitions, multi-character interactions, and complex camera trajectories, randomly generated with ChatGPT \citep{ChatGPT5.2}. These scenarios each contain 3-6 temporal events. Participants were shown videos alongside their corresponding prompt, with model identity withheld, and asked to rank each video on a scale of 1–5 per criterion. Final scores are computed as the average rank across all participants (30) and scenarios.

\begin{table*}[t]
\centering
\resizebox{\linewidth}{!}{
\begin{tabular}{lccccc}
\toprule
\textbf{Metric} & \textbf{Storyboard} & \textbf{Kling 2.6} & \textbf{Veo 3.1} & \textbf{Wan 2.2} & \textbf{Wan 2.2 + Prompt Relay} \\
\midrule
Temporal Prompt Alignment (\(\downarrow\))     & 4.67 & 1.30 & 3.93 & 4.00 & \textbf{1.10} \\
Transition Naturalness (\(\downarrow\)) & 4.60 & 4.43 & 1.30 & 3.50 & \textbf{1.17} \\
Visual Quality (\(\downarrow\))         & 3.67 & 2.50 & \textbf{2.0} & 4.00 & 2.83 \\
\bottomrule
\end{tabular}
}
\caption{Human preference scores for multi-event video generation. (lower values indicate better rankings)}
\label{tab:human_eval}
\end{table*}

\subsection{Results}
As shown in Table. \ref{tab:human_eval}, Prompt Relay consistently outperforms baseline approaches in temporal alignment and transition naturalness. Notably Wan 2.2 with Prompt Relay consistently exhibits stronger visual quality compared to the baseline Wan 2.2. This is likely because Prompt Relay’s attention routing mechanism suppresses attention between queries in a particular temporal segment and prompts belonging to other segments. By reducing unnecessary competition in the cross-attention space, the model can allocate attention more effectively to the active semantic concepts, resulting in clearer visual structure, improved temporal alignment, and more stable generation. However, Kling 2.6 and Veo 3.1 still achieve higher visual quality overall, 
indicating that visual fidelity remains partially bounded by the capacity of 
the underlying backbone model.

\section{Limitations}



Since each temporal segment attends primarily to its corresponding local prompt, persistent visual elements such as characters, objects, or scene style are not explicitly shared across segments. 
If these elements are described inconsistently across local prompts, their appearance 
may drift over time. We found that we can fully mitigate this by incorporating a global prompt 
that provides shared context and anchors persistent elements across multiple segments.

\section{Conclusion}

We present Prompt Relay, an inference-time, plug-and-play method for multi-event video generation with fine-grained temporal control. We also show that our method improves visual quality over the backbone model. We view our work as a pivotal step towards movie-grade, controllable video synthesis.

\section*{Acknowledgments}

This research is supported by cash and in-kind funding from NTU S-Lab and industry partner(s). This study is also supported by the Ministry of Education, Singapore, under its MOE AcRF Tier 2 (MOE-T2EP20223-0002).